\documentclass[10pt,twocolumn,letterpaper]{article}

\usepackage{wacv}
\usepackage{times}
\usepackage{epsfig}
\usepackage{graphicx}
\usepackage{amsmath}
\usepackage{amssymb}

\usepackage[pagebackref=true,breaklinks=true,letterpaper=true,colorlinks,bookmarks=false]{hyperref}

\wacvfinalcopy 

\ifwacvfinal\fi
\begin{document}

\title{Key-Pose Prediction in Cyclic Human Motion}

\author{Dan Zecha \hspace{2cm} Rainer Lienhart \\
Multimedia Computing and Computer Vision Lab, University of Augsburg\\
{\tt\small [dan.zecha|rainer.lienhart]@informatik.uni-augsburg.de}
}

\maketitle
\ifwacvfinal\thispagestyle{empty}\fi

\begin{abstract}
In this paper we study the problem of estimating inner-cyclic time intervals within repetitive motion sequences of top-class swimmers in a swimming channel.
Interval limits are given by temporal occurrences of key-poses, i.e. distinctive postures of the body. A key-pose is defined by means of only one or two specific features of the complete posture. It is often difficult to detect such subtle features directly. We therefore propose the following method: Given that we observe the swimmer from the side, we build a pictorial structure of poselets to robustly identify random support poses within the regular motion of a swimmer. We formulate a maximum likelihood model which predicts a key-pose given the occurrences of multiple support poses within one stroke. The maximum likelihood can be extended with prior knowledge about the temporal location of a key-pose in order to improve the prediction recall. We experimentally show that our models reliably and robustly detect key-poses with a high precision and that their performance can be improved by extending the framework with additional camera views.
\end{abstract}
\section{Introduction}
\begin{figure*}[t]
\begin{center}
\includegraphics[height=60mm]{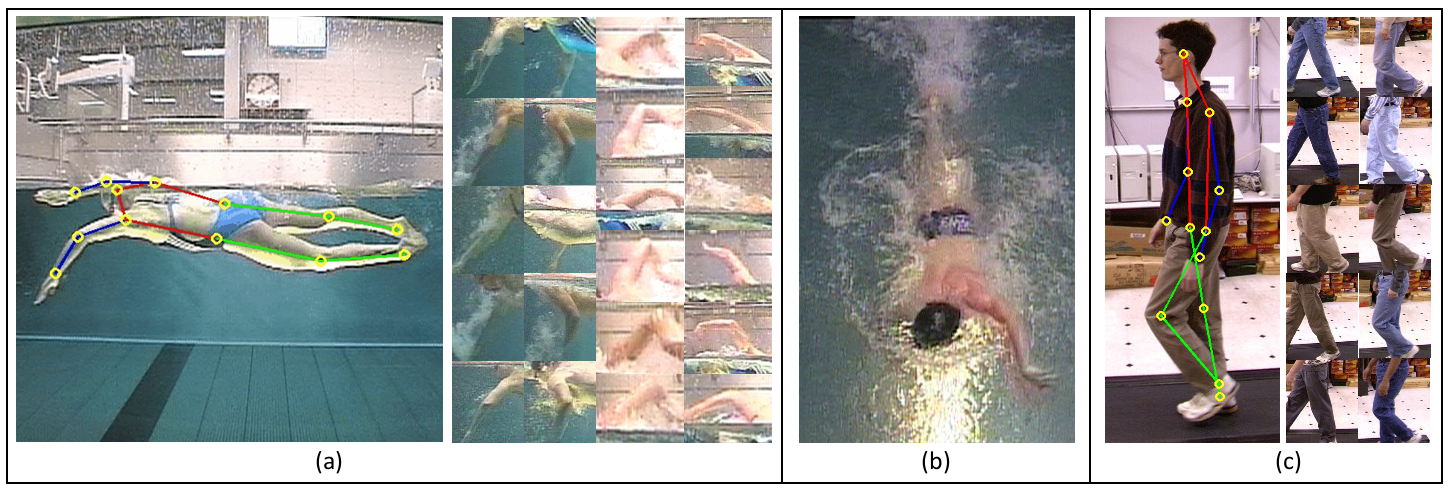}
\end{center}
\caption{(a) A swimmer in a swimming channel (left). Arm configuration patches form clusters for poselets (right). (b) Above view on the swimming channel. (c) Example from the CMU motion of body database with poselet cluster examples. }
\label{swimmer01}
\end{figure*}

In this work we describe an application for top-class competitive sports, taking into account recent developments in pose and motion estimation as well as time series analysis. 
Consider the following real-life application of a camera system for evaluating the technique of swimmers:
In the field of competitive swimming, a quantitative evaluation is highly desirable to supplement the typical qualitative analysis. Therefore, an athlete swims in a swimming channel, a small pool where the water can be accelerated to constantly flow from one end to the other. The pool has at least one glass wall and is monitored with multiple cameras from different angles. The athlete then performs regular swimming motions in one of the four swimming styles, namely freestyle, breaststroke, backstroke or butterfly, while being filmed by one or more cameras. Figure \ref{swimmer01}(a) depicts an example snapshot from one camera for this setup.
The video footage is evaluated afterwards by an experts who for instance annotates certain poses, single joints and other variables of interest. Desired kinematic parameters are inferred from this manual evaluation in a last step.
However, quantitative (manual) evaluations are very time consuming and therefore only used in very few individual cases.
The proposed solution focuses on determining all data necessary to automatically derive desired kinematic parameters of a top-level swimmer in a swimming channel, that is, we would like to retrieve the stroke frequency and inner-cyclic intervals. 
Concluding kinematic parameters from these intervals and recommending actions for improving the technique is not part of our approach; this is the job of a professional coach and falls into the field of training sciences. Hence, we look at this problem solely from a computer vision standpoint.
The problem of determining all aforementioned parameters can be reduced to the following: Given a stream of image frames, we would like to identify poses of special interest, which we label as \textit{key-poses}. In general, \textit{a key-pose is defined by a human expert based on arbitrary features of the pose}. A feature could for instance be the position or angle of the upper arm in the image. 
Detecting such pose features directly in a video stream is quite challenging due to heavy noise and (self-) occlusion. 
We therefore assume that the frames of interest, i.e. the ones showing a key-pose, are ``hidden'' as we cannot detect them directly. A cyclic motion however has a predetermined structure. Hence, we interpolate the occurrence of a key-pose based on points in the cycle that can be detected reliably.   

Therefore, we propose the following method: A database of joint - annotated images of swimmers is temporally clustered and a detector is trained for each cluster. The term temporally refers to a central property of each cluster, which should only contain image patches from a part of poses that appear closely within a small window of time within a swimming cycle.
Multiple detectors are joined into a star-shaped pictorial structure, which outputs a frame descriptor for each image based on spatially restricted max pooling of each part. We aggregate all frame descriptors to form a set of time series in order to decide whether a detector is activated. Based on these activations, the occurrence of a key-pose is estimated by averaging over ``good'' detector signals. Finally, we show that our approach is not limited to the analysis of athletes and apply it to a human gait dataset.

\textbf{Preliminary Definitions.}
Before we dive into the model formulation, we would like to define some commonly used terms in this work.
The smallest unit within a repetitive motion is one \textit{cycle} or \textit{stroke}, defined by the time that passes between the appearance of a specific body pose and its earliest reappearance. 
Freestyle swimming (like walking or running) is an anti-symmetrical cyclic motion: Every pose of the left body half occurs approximately half a cycle later on the other side of the body. This observation is important as a detector based on gradients is not able to reliably distinguish anti-symmetrical poses given that we observe the person from the side.

\section{Related Work}
Part based models have played a huge role in the fields of object detection and (human) pose estimation within the last years. Based on the fundamental work Fischler and Elschlager~\cite{Fischler:1973:RMP:1309264.1309318}, these models represent an object through multiple parts which are connected via deformation terms, often visualized as springs, allowing for matching them in a flexible configuration. Various manifestations of this basic notion have been developed through the years, kicked of by Felzenszwalb et al~\cite{lsvm-pami} with their deformable part models for object detection. Different refinements have been proposed specifically for human pose estimation, e.g. by enriching a model with additional parts to compensate for the flexibility of the human body~\cite{yang2011articulated} or by allowing rotation of parts~\cite{Andriluka:2009}.

While effective implementations of part detectors have been proposed for characteristic body parts like head and torso, part templates for extremities are usually weak. This issue has been addressed by~\cite{pishchulin13iccv}, who argue that person and body part templates should be pose specific rather than generally trained. Bourdev et al~\cite{BourdevMalikICCV09} also follow this notion by proposing the concept of poselets, generic rigid part detectors based on Histograms of Oriented Gradients (HoG)~\cite{DT05} as a generalization of specific body part detectors. Poselets lift the spatial limitation of parts being connected to an actual body part and encode generic parts of the body. Gkioxari et al~\cite{Gkioxari2013} recently utilized poselets for training discriminative classifiers to specifically differentiate between arm configurations of a person.

In the context of key-frame selection in videos, poselets have been used for human activity recognition. \cite{Raptis2013} proposed a framework based on poselet activations for selecting key-frames that represent key-states in an action sequence. An additional classifier trained on pairwise terms for all activations then decides if a specific action sequence occurred. Carson et al~\cite{Carlsson01actionrecognition} select action specific postures by matching shape information from individual frames in order to recognize specific tennis strokes in game footage. 

The analysis of human gait probably plays the biggest role in the field of periodic motion research. A big focus lies on the identification of a person via his/her intrinsic gait signature, for example by determining and tracking the body shape \cite{1384807} or through fusion of multiple gait cycles \cite{5334059}. More general approaches strive to recover the human body pose \cite{1613028} in order to retrieve a full set of gait parameters. Periodic motion in images was examined by Cutler \& Davis \cite{Cutler99robustreal-time}, who use self similarity and frequency transformations to obtain the frequency of the periodic motion of human and animal gait.

Most work researching the tracking of people in aquatic environments has focused on drowning detection~\cite{EngTYW08}, localization of athletes in swimming competitions~\cite{longsha} and motion analysis for video based swimming style recognition~\cite{tong}. A Kalman filter framework is presented in~\cite{greifswimmer} to explicitly model the kinematics of cyclic motions of humans in order to estimate the joint trajectories of backstroke swimmers. Ries et al~\cite{riesswimmer} use Gaussian features for detecting a specific pose of a swimmer in a pool with the intention of initializing his/her pose. The method closest to our approach is presented in~\cite{zechaswimmer}, who divide swimming cycles into intervals and train object detectors for each interval. The stroke rate is computed by counting the occurrences of the intervals. However, they show that arbitrary poses cannot be detected with their approach.

\section{Method}
For deducing desired key-poses from a predictable or repetitive motion, we build a two staged system:
Firstly, a pictorial structure of poselets is trained in order to extract a descriptor for each frame in the video.
Secondly, we aggregate all frame descriptors to time series and define a maximum likelihood estimator for good poselet signals in order to predict the occurrence of a key-pose.

\subsection{Poselet Training}
We build our system on localizable parts of the human pose, initially introduced as poselets by Bourdev et al.\cite{BourdevMalikICCV09}. The original work defines poselets as rigid linear filters, based on Histograms of Oriented Gradients (HoG, \cite{DT05}) features. Each filter is trained from a set of example image patches that are close in configuration space. The patches are transformed into feature space and a linear SVM is trained for them. For evaluation, the resulting dense linear filter is  cross-correlated with a feature grid/pyramid, yielding a score for every placement of the filter.

Our approach depends heavily on precisely trained, distinguishable poselets.  We achieve this by extracting patches of characteristic parts of the motion, e.g. patches of the limbs, from all images. The underlying groups of joints, called configurations, are clustered by a k-means algorithm. From each resulting group of patches a linear SVM is trained.

A simplifying assumption often made in the context of poselets is that the representation of the part does not change with part rotation. While other approaches \cite{Gkioxari2013} strive to find the best possible  transformation between different configurations by rotating, reflecting, translating and rescaling body configurations, we would like to \textit{extend the notion of a poselet as a representation of a part of the pose and additionally a small time window within a repetitive motion}. Hence, we develop the following distance function for the clustering algorithm which assures that the configurations are not rotated.

\textbf{Dataset.} Our dataset for training the poselets is build on video footage of freestyle swimmers.
We annotated 1200 images with complete configurations, i.e. a total of 13 joints per athlete. In case of (self) occlusion, we averaged between joint location estimates of different annotators. The images cover different athletes, 3 male and 5 female, performing overall 20 strokes. We tried to cover most obvious variables that influence the configuration space and image quality, e.g. different genders, body heights, physiques, illumination and water flow velocities (between $1ms^{-1}$ and $1.75ms^{-1}$). All swimmers are filmed by one camera trough the side wall of the swimming channel, depicting their left body side. The camera films with a resolution of $720 \times 576$ pixels at $50i$. From these images, groups of ``sub-configurations'' are extracted, e.g. arm-configurations (shoulder, elbow and wrist of the same arm) or leg configurations (hips and knees). The clustering algorithm then groups these sets of joints and each cluster forms the foundation for training a linear SVM (poselet).

\textbf{Temporal Poselet Clustering.}
Let $\boldsymbol{A}=(\boldsymbol{a}_1,\cdots, \boldsymbol{a}_n)^T \in \mathbb{R}^{n,2}$ and $\boldsymbol{B}=(\boldsymbol{b}_1,\cdots, \boldsymbol{b}_n)^T \in \mathbb{R}^{n,2}$ be two configurations of $n$ joints, where each $\boldsymbol{a}_i $,$\boldsymbol{b}_i \in \mathbb{R}^2$ $(i = 1, \cdots, n)$ denotes a 2 dimensional location $(x,y)$ of one joint. We wish to find a transformation that moves $\boldsymbol{B}$ to $\boldsymbol{A}$ so that the euclidean square norm is minimized, i.e.
\begin{equation}
\label{min_transform1}
d(\boldsymbol{A}, \boldsymbol{B}) = \min_{s,c} \sum_{i=1}^{n}\parallel \boldsymbol{a_i} - s\boldsymbol{b_i} + \boldsymbol{c} \parallel_2^2.
\end{equation}
Thus the configuration $\boldsymbol{B}$ is translated via $\boldsymbol{c}$ and resized with a scaling factor $s$. This formulation closely resembles the Procrustes optimization problem~\cite{seber}, with the important difference that we do not allow for $\boldsymbol{B}$ to be rotated and reflected by a linear transformation. This will assure that a configuration from any point within 	the cyclic motion is not reflected or rotated onto a configuration that is not temporally close. 
One solution for equation \ref{min_transform1} is given by
\begin{equation}
\label{final_distance}
d(\boldsymbol{\bar{A}}, \boldsymbol{\bar{B}}) = \frac{\sum_{i=1}^{n}\parallel \boldsymbol{\bar{a}}_i -  \max(0,tr[\boldsymbol{\bar{A}}\boldsymbol{\bar{B}}^T]/tr[\boldsymbol{\bar{B}} \boldsymbol{\bar{B}}^T] \boldsymbol{\bar{b}}_i) \parallel _2^2}{ tr\left[ \boldsymbol{\bar{A}} \boldsymbol{\bar{A}}^T\right]},
\end{equation}
where $\boldsymbol{\bar{A}}$, $\boldsymbol{\bar{B}}$ and $\boldsymbol{\bar{a}_i}$, $\boldsymbol{\bar{b}_i} $ are mean corrected matrices and vectors respectively.
The max operator in equation \ref{final_distance} constraints $s$ to be greater or equal than zero. A negative scaling factor is equivalent to a point reflection at the origin of the coordinate system. As we stated that any kind of reflection is unwanted behaviour, we force the distance function to the closest optimal solution w.r.t. the constraint. The denominator of equation \ref{final_distance} was suggested by Sibson \cite{sibson} and standardizes the distances between different pairs of configurations with the intention of making them comparable. 

A k-means clustering is applied using distance function in \ref{final_distance}, yielding groups of image patches that occur temporally close within a cycle. Hence, each cluster represents a part of an athlete's body within a small time frame. All image patches in a cluster are transformed into dense HoG grids and a linear SVM is trained on them, yielding one poselet per cluster.

\subsection{Frame Descriptor}
\label{frame_descriptor}
Recently poselets have been used in human detection and pose estimation \cite{Raptis2013}. 

Given a set of poselets trained from different temporally close configurations, we would like to pick up the notion of a poselet activation vector \cite{BourdevPoseletsECCV10}. Instead of simply maximizing over the cross-correlations of a poselets at all positions and scales of a feature pyramid, we add a spatial bias to our model: A poselet is only evaluated in a region relative to the location of an athlete, which is determined by an additional detector trained for the complete configuration of a swimmer. This guarantees that we do not search for a part of the human body if there is no person present in the image or close to its position. We take advantage of the temporal component of our system in order to decide if a poselet is activated by observing the score of a detector over time and deciding that it is activated if its score is a locally maximal.

\textbf{Mixture of poselets.} Let $P_i=(F_i, w_i, h_i)$ $(i= 1, \dots, n)$ be a triple describing one of $n$ poselets $F_i$ with size $w_i \times h_i$. The score of a poselet at position p in a feature pyramid is computed via the cross correlation
\begin{equation}
\text{score}(F_i, p) = F_i \star \Phi(p)
\end{equation}
of $F_i$ with the underlying subwindow $\Phi(p)$ in the pyramid. A position $p = (x, y, s, w, h)^T$ is defined by two coordinates x and y, a scale $s$ of a pyramid level and the size of the subwindow $w \times h$, which equals the size of the poselet. 

Multiple poselets are combined into a mixture $M = (P_0, P_1, \dots, P_{n})$. Similar to \cite{DT05}, we train a poselet $F_0$ for the whole configuration of an athlete and use it to retrieve an initial hypotesis $p_0$ for the placement of the athlete, where
\begin{equation}
\label{bestposition}
p_0 = \underset{p}{\text{argmax}} \text{  score}(F_0, p).
\end{equation}
This best detection $p_0$ is projected to the original image size through $\widehat{p}_0 = p_0 s_0^{-1} = (xs_0^{-1}, y s_0^{-1}, 1, w_0s_0^{-1}, h_0s_0^{-1})^T = (\widehat{x}_0, \widehat{y}_0, 1, \widehat{w}_0, \widehat{h}_0)^T$. With this root hypothesis, we retrieve a score for each poselet $P_i$ in the mixture by maximizing over
\begin{equation}
\label{indicator}
s_{f,F_i} = \max_{p_k \in R} \text{score}\left(F_i, p_k \right),
\end{equation}
where
\begin{equation}
R = \left\{ p_k \bigg{|} \sqrt{z^T \left( s_k^2 \Sigma \right)^{-1}  z } < \gamma\
\right\}
\end{equation}
with $z \equiv (x_k, y_k)^T - s_k(( \widehat{x}_0, \widehat{y}_0)^T+\mu_i)$.
The set $R$ restricts the position of a poselet by means of the Mahalanobis distance. All position elements in $R$ lie within an elliptic region defined by the covariance matrix $\Sigma_i \in \mathbb{R}^{2\times2}$. It is centered around the position of the root model $F_0$ plus an offset $\mu_i \in \mathbb{R}^2$ of the poselet relative to the root. The size of this region is restricted by $\gamma$ and empirically set to $\gamma = 3$. Both $\mu$ and $\Sigma$ can be estimated directly from the training data by fitting a normal distribution on pairwise offsets between the root model and a poselet. The final output of our model is a frame descriptor $s_f$ of spatially limited poselet scores, where
\begin{equation}
\boldsymbol{s}_f = 
(s_{f,F_1}, \hdots , s_{f,F_n})^T
\end{equation}
for frame $f$. Note that the scores of each poselet are not thresholded in any way to determine if it is activated. We determine if a poselet is active by examining the poselet scores over time instead in the next section.

The model formulation above closely resembles a star shaped pictorial structure of poselets, where all parts are connected via deformation features $\mu$ and $\Sigma$ with a root model. A popular related approach, initially developed by Felzenszwalb et al \cite{lsvm-pami} and expanded by many others \cite{yang2011articulated, pishchulin13iccv, Andriluka:2009}, is called a deformable part model for object detection. Similar to these models, we can solve the problem described in \ref{bestposition} and \ref{indicator} efficiently using dynamic programming. 

Note that the root model $P_0$ (and also all part models) do not necessarily have to be poselets; equivalently, they may be replaced by more sophisticated or complex models without loss of generality of this approach.

\subsection{Key-pose estimation}
\label{MLestimate}
In order to estimate the (regular) occurrence of a key-pose in a video, we firstly post-process aggregated frame descriptors and define a measure of goodness for a time series of max-pooled poselet scores based on self similarity of the series. Secondly, we describe a maximum likelihood estimator for predicting a key-pose.

The frame descriptors from section \ref{frame_descriptor} are aggregated in a matrix $S = (\boldsymbol{s}_1,\cdots,\boldsymbol{s}_T)$ for the $T$ frames of a video.
As we trained the mixture model for temporally distinctive detectors, each poselet vividly acts like a sensor measuring the presence or absence of a body part. If the body part that the poselet was trained for is present in a frame at the location specified by the deformation variables, the score of the poselet should be high. If the athlete continues his movement, the position or representation of a body part changes and the poselet score should decrease. The underlying assumption here is that the poselet ``works'', i.e. that the observed score really resembles the image content. This is of course not the case for all poselets: While some configurations are simply not suited to be represented and reliably detected by a dense linear filter based on HoG features, other configurations are not present for a specific swimmer. 

Recall that the poselets for our mixture model are trained on images depicting the swimmer from the side. We found that dense HoG templates trained for arms are not able to distinguishing between left and right arms. As a consequence, we observe that the score of a working poselet has two peaks within one stroke for anti-symmetrical swimming styles. This is not a problem in general if we adjust all evaluation criteria accordingly.
Before we assess the quality of a poselet time series, we post process all series as follows.

\textbf{Time Series Post Processing.} 
Let $S_i = (s_{1,i}, \cdots, s_{T,i})$ be the time series describing the score of poselet $P_i$ over time.
In order to compensate for the noisy output of linear HoG filters, we smooth each time series with a Gaussian filter. 
The activation of a poselet is then given by locations of local maxima $m_i \in M_i = \{m_{i,t}\in \mathbb{N}^+\}$ in $S_i$, i.e. $M$ holds the frame numbers where a poselet detection is locally maximal. For a good poselet, the distance $m_{i,t+1}-m_{i,t-1} $ for $1< t < |M|$ equals the time of one complete stroke for all anti-symmetrical swimming styles. For breaststroke and butterfly, this time period is equivalently given by $m_{i,t+1}-m_{i,t} $, as a poselet only has one score peak within a stroke. We finally build a histogram for all stroke intervals within a sliding window of $S_i$ in order to determine the main stoke frequency $f_{stroke}$ for the swimmer and iteratively discard obvious false detections in $S_i$ by greedily deleting occurrences in $M_i$ that produce frequencies much smaller than $f_{stroke}$. All intervals $[m_{i,t-1},m_{i,t+1}]$ are called regular, iff $m_{i,t+1}-m_{i,t-1}-f_{stroke}<\lambda$ holds for a small $\lambda$ (e.g. $\lambda=0.1\cdot f_{stroke}$).

Finally, we sort all poselet activation series by their ``goodness'': By computing the error between adjacent groups of activations within a poselet series, we get a small error if the series is very regular (i.e. it is well suited for the prediction step and therefore a good series) and a larger error if the series is irregular due to additional, missed or false detections. These series will introduce errors in our key-pose predictions and are therefore ill-suited.

\textbf{Key-pose prediction.}
The final step in our framework tries to find the best estimate of an occurrence of a key-pose $g$ given that we observe the activations of $n$ spatially constricted poselet activations $m_i$, i.e.
\begin{equation}
\label{maxlikeli}
\widehat{g} = \underset{g}{\text{argmax}} \text{  }p(g|m_1,\cdots,m_n)
\end{equation}
We can rewrite this MAP hypothesis by means of Bayes' theorem, presupposing independence between all poselets, yielding
\begin{equation}
\label{maxlikeli2}
\widehat{g} = 
\underset{g}{\text{argmax}} \text{ } \sum_i^n \log(p(m_i|g)) + \log(p(g)).
\end{equation}
For starters, we assume a uniformly distributed prior and only look into the maximum likelihood of equation \ref{maxlikeli2}.
Vividly, each poselet gives its own estimate of where a key-pose occurs in the time series, modeled by the likelihood $p(m_i|g)$ of poselet i. The final position is then ``averaged'' between different estimates.
The likelihoods in equation \ref{maxlikeli2} can be modeled by applying our pictorial structure of poselets on a set of videos where the ground truth frame numbers for a key-pose were annotated by a human expert. 
The resulting time series for all videos are post-processed as described above. The likelihood for poselet i is then modeled by a normal distribution for ground truths relative to poselet activations. We therefore collect regular activation intervals $[m_{i,t-1},m_{i,t+1}]$. The anti-symmetry property of freestyle swimmers yields two ground truth occurrences $g_{t-1}$, $g_t \in \mathbb{N}^+$ (i.e. one for the left arm, one for the right arm) in between them.
The likelihood $p(m_i|g)$ is modeled by a Gaussian $\mathcal{N}(x;\mu_i, \sigma_i)$ fitted to all c, where
\begin{equation}
\label{c_coeff}
c = \frac{g_t-m_{i,t-1}}{m_{i,t+1}-m_{i,t-1}}.
\end{equation}
The denominator normalized the occurrence $g_t$, making c independent of the stroke frequency. \textit{Note that a frame $g_t$ depicting a key-pose has to be given here. As a key-pose is always defined by a human-expert, all key-poses have to be manually annotated in order to train a key-pose prediction model.}

At inference time, we directly compute the ML hypothesis, ignoring all non regular intervals. Each good poselet time series yields regular intervals $[m_{i,t-1},m_{i,t+1}]$ which generate a list of possible occurrence estimates $\mu_{pos}$, that is
\begin{equation}
\mu_{pos} = m_{i,t-1} + \mu_i(m_{i,t+1}-m_{i,t-1})
\end{equation}
and an uncertainty $\sigma_{pos}$ for each occurrence, with
\begin{equation}
\sigma_{pos} = \sigma_i(m_{i,t+1}-m_{i,t-1}).
\end{equation}
Finally, the sum over all individual likelihoods is evaluated within small subwindows around multiple guesses $\mu_{pos}$. Hence, an occurrence k is given by
\begin{equation}
\label{occurrence}
\text{occ}_k = \underset{x}{\text{argmax}} \sum_{\mu_{pos} \in \text{subwindow}} \log(\mathcal{N}(x;\mu_{pos}, \sigma_{pos})).
\end{equation}
We empirically found that good placements for subwindows are given by the locations of local maxima in $(g(\Sigma) \star \sum_{pos} \mathcal{N}(x;\mu_{pos},\sigma_{pos}))$, with $g(\Sigma)$ being a Gaussian smoothing kernel.

The formulation above assumes that the prior probability $p(g)$ is uniformly distributed, thus reducing the MAP hypothesis to an ML estimate. The complete hypothesis however assumes that we have a prior for an initial key-pose frame. For instance, an expert could annotate just one single occurrence of a key-pose for a specific swimmer manually in order to improve the likelihood estimate with previous knowledge $p(g)$. We could then propagate this single annotation to all other cycles by building a (probably incomplete) model based on the idea presented in Equation \ref{c_coeff}, setting the standard deviation $\sigma_{pos}$ to a fixed value. We will show the effects of ignoring or setting the prior guess in the experimental section.

\begin{figure*}[!t]
\begin{center}
\includegraphics[height=38mm]{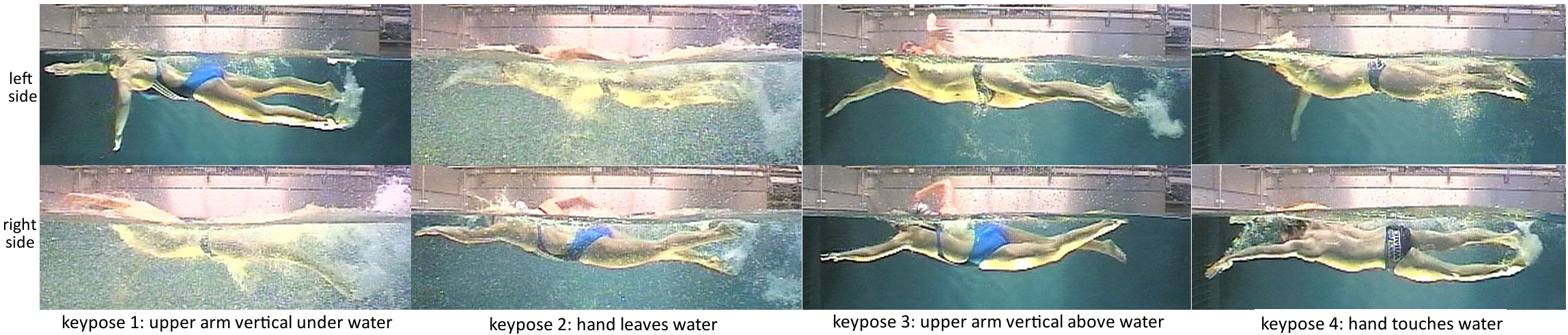}
\end{center}
\caption{Four different key-poses for a freestyle swimmer. Key-poses occur on the left and on the right side of the body.}
\label{allkeyposes}
\end{figure*}
\section{Experimental Results}
We validate and discuss the performance of the proposed likelihood estimators on a set of 30 swimmer videos (720x576@50i) covering different freestyle swimmers (6 male, 8 female, ages 15 to 25, different body sizes) in a swimming channel with slowly increasing water flow velocities (minimal velocity $1ms^{-1}$, increase of maximal $0.3ms^{-1}$, maximal velocity $1.75ms^{-1}$). The videos show the swimmers from the side. An expert annotated all frames that depict one of four key-poses (Figure \ref{allkeyposes}, overall $4 \cdot 424$ occurrences). 

We trained a 16 part pictorial structure model of poselets (1 root, 15 arm poselets) from 1200 distinctive images. 
We do not directly evaluate the performance of each poselet but instead show their efficiency in the following analysis of the activation sequences. 
The estimation of key-pose occurrences is evaluated in a 30-fold leave-one-out cross-validation, where we extract ML estimators for all combinations of 29 videos and evaluate their performance on the remaining video. 

\textbf{Performance measures.}
In general, we distinguish different types of detections: If our detectors estimate an occurrence within 10 half-frames of a ground truth annotation, the prediction is counted as a true positive ($TP$) detection and as a false positive ($FP$) otherwise. A ground-truth frame without a prediction is a false negative ($FN$). In the following, we will compare the strokelength-normalized deviation of a prediction from the ground-truth annotation (i.e. percent of deviation from optimum, on the x-axis) with the recall of the system (y-axis), which is defined as $rec=TP/(TP+FN)$. The recall is an indicator of how many key-pose occurrences we have estimated correctly. All recall vs. deviation curves are evaluated at a deviation $x=0.03$ 
 This threshold frames the error human annotators make when annotating the ground truth. A deviation of 3\% reflects a slack between $\pm 1$ to $\pm 2$ half frames on average (for different videos with different water flow velocities). For each recall vs. deviation graph, we will also compute the precision of the model, which is defined as $prec = TP/(TP+FP)$ and an indicator for how many false guesses we made. 

\begin{figure*}[t]
\begin{center}
\includegraphics[height=39mm]{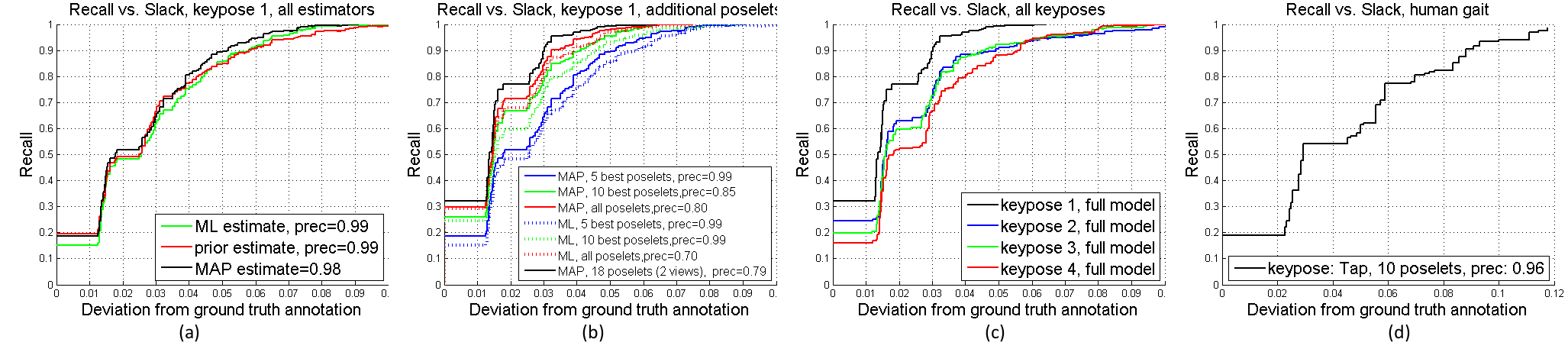}
\end{center}
\caption{ Recall of our system for different slacks. (a) Comparison of different estimation methods (ML, prior, MAP) for keypose 1. (b) Recall for predictions from different numbers of support poses. (c) Performance comparison for different key-poses. (d) Recall for our system on the CMU MoBo dataset \cite{CMU} for slow walking.
}
\label{plot01}
\end{figure*}

\textbf{ML, prior and MAP predictions.}
We evaluate the maximum likelihood model described in section \ref{MLestimate} exemplary for key-pose 1. Figure \ref{plot01}(a) visualizes the recall relative to the deviation from ground truth annotated frames. The model computes the goodness of each poselet activation time series. By using the time series of the 5 best performing poselets, it achieves a recall of $61\%$ at a deviation of $3\%$ and a precision of $0.99$, which means that we made a small number of false estimations. False positives are predicted because it is not guaranteed that we always find 5 good performing poselets a time.
We furthermore evaluated a prior estimate and the complete MAP prediction from section \ref{MLestimate}. A model for the prior can be extracted from just a single expert annotation for each key-pose in same manner as the ML model. This model might be incomplete though, as we can't find regular stroke intervals for all poselets framing this one ground truth annotation. Also, we are not able to set an uncertainty $\sigma_{pos}$ for just one example, although we can assume that the annotation is probably very good for this swimmer and set a fixed small value (empirically: $\sigma_{pos} = 0.04 \cdot f_{stroke}$). We can apply this model alone or join it with the maximum likelihood to complete the MAP estimator. While the prior alone performs unsurprisingly well for all videos and a little better than the ML, it is slightly surpassed by the outcome of the complete MAP estimate (Figure \ref{plot01}(a)).

\textbf{Performance improvements.}
In order to enhance the performance of our system, we continually added additional ``bad'' poselet time series to our MAP estimates (Figure \ref{plot01}(b)). While the overall performance clearly improves the predictions up to a recall rate of $0.85$ if we use all poselets (even disadvantageous ones) from the model, the precision drops down to $0.80$ as an unwanted side effect . We found that most bad poselet time series, even the bad ones, contain at least some valid stroke intervals which improve the performance. However, they of course contain a lot of regular stroke intervals which do not fit the regularity of the complete signal; these intervals will produce a lot of false predictions. Surprisingly, while the ML estimator always performs a little worse compared with the MAP estimator, it keeps its good precision longer (see precision comparison for green graphs). We found that the precision can be improved by applying the same heuristics used for post-processing single activation series in section \ref{MLestimate} to the key-pose estimation series. Additionally, we condition the sliding window approach for averaging between different single poselet predictions (Equation \ref{occurrence}) so that a minimum of two predictions have to be within the subwindow. Both heuristics effectively kill nearly all remaining false positives in any key-pose occurrence series, \textit{leaving us with an acceptable precision of $prec >0.98$}.

\textbf{Additional camera view.}
The insight about improving estimates with additional, if possible good poselet time series, inspired the following experiment: We trained an additional 7-part mixture of poselets for a second camera view (Figure \ref{swimmer01}(b)) that captures each swimmer from above the swimming channel. While this camera does not monitor any movement below the water line because of the turbulent water surface, it detects the swimmer and any arm movement above the water line very well. Note that this model behaves like a model trained for a symmetrical swimming style: poselet activations only occur once a cycle (instead of two times for an anti-symmetrical swimmer observed from the side) because each arm is detected by its own set of poselets. While interpolating key-poses for this type of model is a bit easier (time series of anti-symmetrical styles can be interpreted as two independent, superimposed event signals, one for each body side), we want to join their time series with our side-view results. Therefore, we condense pairs of two poselet time series from the above view, so that the same semantic postures of the arms on both sides of the swimmer form a new series. For the 7-part model, we hence get an additional 3 time series, which we join with the other 15 time series which we extracted from the first view. As a  result the recall improves another $4.5\%$ to an overall of $0.89$, with no significant change for the precision (drop of 0.01). This final prediction result is depicted in Figure \ref{plot01}(b) (black graph).

\textbf{Comparison between different key-poses.}
Our approach is of course not limited to only one key-pose for a cyclic motion. In fact, if we would like to compute inner-cyclic intervals, we need more than one measuring point within a stroke. Therefore, an expert annotated four different key-poses (depicted in Figure \ref{allkeyposes}) and we trained one combination of poselet+MAP model for each pose. Figure \ref{plot01}(c) shows the deviation from ground truth frames for these four key-poses. We observe that the predictions for some key-poses are better than others. This behavior can be explained by the fact that cyclic motion is of course not linear (or constant) in its acceleration and velocity. We found that our model has some difficulties in precisely detecting key-poses in intervals where the velocity of an arm is very small over a timespan of 10 to 20 frames. Additionally, we found that annotating the ground truth for worse performing estimators was always more difficult due to heavy occlusion and image noise. 

\textbf{Human Gait Dataset.}
Although we developed our models originally for swimmers in swimming channels, they are not bound to this application. In order to prove that our approach can be used for any regular cyclic motion, we carried out another experiment on the CMU Motion of Body database \cite{CMU}, which is one of the better known human gait datasets. We trained a 10 part pictorial structure for leg configurations of 10 slowly walking persons depicted from the left side (Figure \ref{swimmer01}(c)). The poselet model is completed with a ML estimator which was trained for the key-pose where either the left or the right heel of the test person touches the ground again (end of swing phase, beginning of stance phase). The model was then evaluated on a different set of 10 video depicting slowly walking test persons with the same evaluation criterion as for the swimmers. The result is depicted in Figure \ref{plot01}(d). As the ``walking frequency'' is usually higher than the stroke-frequency of a swimmer, an acceptable deviation of $\pm 2$ frames is equivalent to $8\%$. Within this range, $80\%$ ``tap-events'' are classified correctly with a precision of $0.96$. 

\textbf{Comparison to other approaches.}
Comparing our approach with other system is difficult: To our knowledge, the detection of key-poses in repetitive motion of athletes with the intension of retrieving the stroke frequency and inner cyclic intervals has not been researched directly. The only comparable approach \cite{zechaswimmer} trains complete object detectors on segments of a swimming cycle. The stroke rate is then extracted by counting the occurrences of an interval in a video. However, they show that their system is not able to detect arbitrary poses.

\section{Conclusion}
We presented a system for estimating the occurrences of key-poses of top-level swimmers in a swimming channel. 
We showed that while it might be difficult to detect a key-pose feature directly, we can estimate the occurrence of such a pose with a high reliability. Future work will focus on extending the approach to other swimming styles and further sports. An additional interesting question is how our models can be transfered to work on arbitrary (competitive) swimmers in swimming pools in general.

{\small
\bibliographystyle{ieee}
\bibliography{egpaper_final}
}

\end{document}